\documentclass[a4paper, 10pt, conference]{ieeeconf}      % Use this line for a4
\usepackage{FG2021}

\FGfinalcopy % *** Uncomment this line for the final submission

\IEEEoverridecommandlockouts                              % This command is only
\usepackage{AbdAlmageed_style}
\usepackage{Jiazhi_style}
\usepackage{subcaption}
\usepackage{subfiles}
\usepackage{color,soul}
\usepackage{xcolor}
\usepackage{multirow}
\usepackage{bm}
\usepackage{makecell}
\usepackage[accsupp]{axessibility}

\def\FGPaperID{211} % *** Enter the FG2021 Paper ID here

\title{\LARGE \bf
Information-Theoretic Bias Assessment Of Learned Representations Of Pretrained Face Recognition
}

\author{\parbox{16cm}{\centering
    {\large Jiazhi Li, and Wael Abd-Almageed}\\
    {\normalsize
    University of Southern California, Information Sciences Institute, Marina del Rey, CA, USA\\}}
}

\begin{document}

%%%%%%%%%%%%%%
% COPYRIGHT NOTICE - Uncomment correct version below
%
% The notices are from the FG 2021 LOA 
%
% Active is the "Others" option - see Case #4 in the instructions posted at: http://iab-rubric.org/fg2021
%
%%%%%%%%%%%%%%

% Case #1: For papers in which all authors are employed by the US government, the copyright notice is: 
%\IEEEoverridecommandlockouts\pubid{\makebox[\columnwidth]{U.S. Government work not protected by U.S. copyright \hfill}
%\hspace{\columnsep}\makebox[\columnwidth]{ }}

% Case #2: For papers in which all authors are employed by a Crown government (UK, Canada, and Australia), the copyright notice is:
%\IEEEoverridecommandlockouts\pubid{\makebox[\columnwidth]{978-1-6654-3176-7/21/\$31.00~\copyright{}2021 Crown \hfill}
%\hspace{\columnsep}\makebox[\columnwidth]{ }}

% Case #3: For papers in which all authors are employed by the European Union, the copyright notice is:
%\IEEEoverridecommandlockouts\pubid{\makebox[\columnwidth]{978-1-6654-3176-7/21/\$31.00~\copyright{}2021 European Union \hfill}
%\hspace{\columnsep}\makebox[\columnwidth]{ }}

% Case #4: For all other papers the copyright notice is:
\IEEEoverridecommandlockouts\pubid{\makebox[\columnwidth]{978-1-6654-3176-7/21/\$31.00~\copyright{}2021 IEEE \hfill}
\hspace{\columnsep}\makebox[\columnwidth]{ }}

\ifFGfinal
\thispagestyle{empty}
\pagestyle{empty}
\else
\author{Anonymous FG2021 submission\\ Paper ID \FGPaperID \\}
\pagestyle{plain}
\fi
\maketitle

%%%%%%%%%%%%%%%%%%%%%%%%%%%%%%%%%%%%%%%%%%%%%%%%%%%%%%%%%%%%%%%%%%%%%%%%%%%%%%%%

%%%%%%%%% ABSTRACT
\begin{abstract}
As equality issues in the use of face recognition have garnered a lot of attention lately, greater efforts have been made to debiased deep learning models to improve fairness to minorities. However, there is still no clear definition nor sufficient analysis for bias assessment metrics. We propose an information-theoretic, independent bias assessment metric to identify degree of bias against protected demographic attributes from learned representations of pretrained facial recognition systems. Our metric differs from other methods that rely on classification accuracy or examine the differences between ground truth and predicted labels of protected attributes predicted using a shallow network. Also, we argue, theoretically and experimentally, that \emph{logits}-level loss is not adequate to explain bias since predictors based on neural networks will always find correlations. Further, we present a synthetic dataset that mitigates the issue of insufficient samples in certain cohorts. Lastly, we establish a benchmark metric by presenting advantages in clear discrimination and small variation comparing with other metrics, and evaluate the performance of different debiased models with the proposed metric.
\end{abstract}

%%%%%%%%% BODY TEXT
\section{Introduction}
The social impact of deep learning has been under scrutiny with the recent rapid developments in many application domains, such as face recognition~\cite{survey} being used in surveillance and security~\cite{surveillance}. One of the major concerns is demographic bias of deep learning systems with respect to \emph{protected attributes}, such as sex~\cite{sex_race_PPB}, race~\cite{race2} or age~\cite{debias_adversarial_Age_sex_race_biasness}. The bias is reflected in the unequal algorithmic accuracy for different demographic groups, such that in facial recognition systems, black females are less likely to be correctly recognized than white males~\cite{sex_race_PPB} and people of color tend to be mistakenly recognized than people of European origin~\cite{bias_surveillance}. Bias issues, hampering generalization for facial recognition systems, are in part due to the end-to-end nature of training deep learning systems, which primarily focuses on minimizing empirical loss in order to maximize recognition accuracy at the expense of encouraging the model to exploit the information from \emph{protected attributes}.

\begin{figure}[t]
\begin{center}
%\fbox{\rule{0pt}{2in} 
%\rule{0.9\linewidth}{0pt}}
   \includegraphics[width=0.83\linewidth]{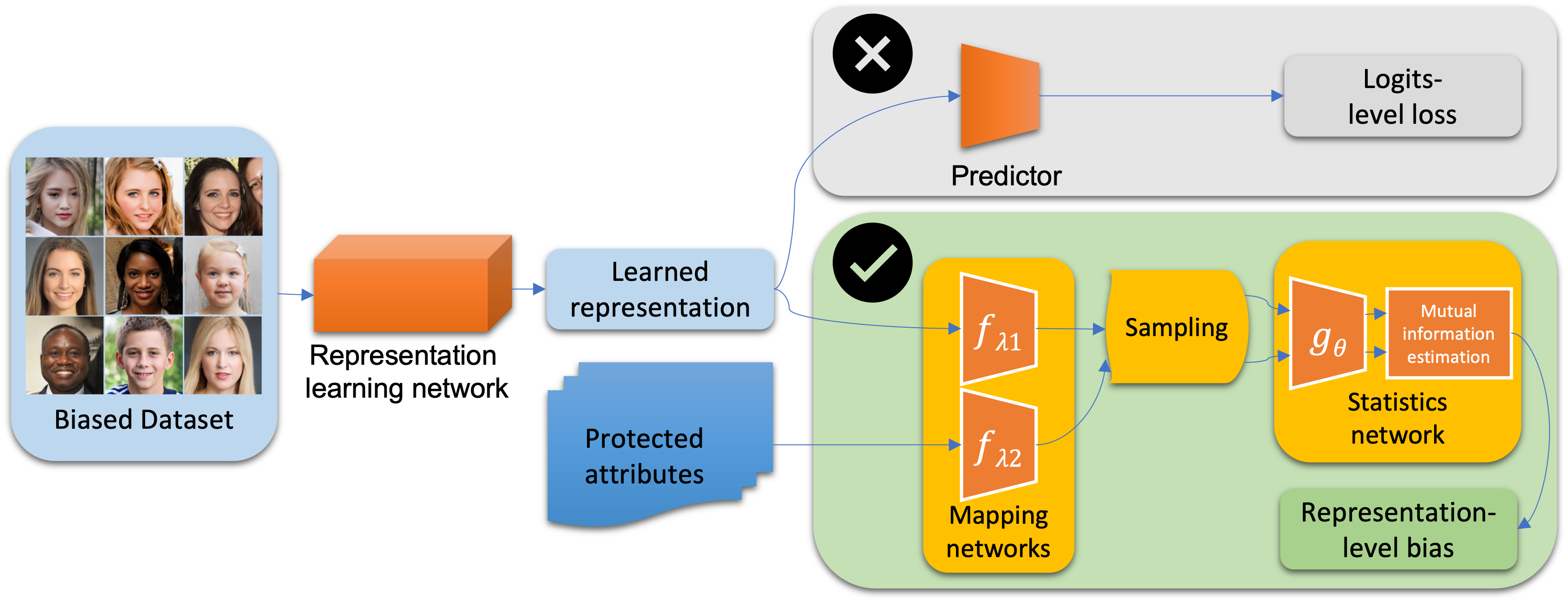}
\end{center}
   \caption{We propose an information-theoretic 
    model to assess the bias involving protected attributes in facial recognition systems. Our approach directly establishes the correlation between learned representation $\mathcal{R}$ and protected attributes $\mathcal{Z}$ with mutual information without relying on any predictors at the \emph{logits} level.}
\label{fig:idea}
\vspace{-0.7cm}
\end{figure}

Mitigating bias, commonly referred to as \emph{debiasing} in face recognition literature, has, therefore, garnered tremendous interest~\cite{debias_resampling_representation_bias_colored_MNIST,sex_race_PPB,debias_domain_independent_training_BA_usage}. Without loss of generality, methods addressing algorithmic bias can be grouped into two families. First, a group of methods that attempt to mitigate bias by improving the diversity and inclusion of training datasets by resampling~\cite{debias_resampling_representation_bias_colored_MNIST} or adding diversity~\cite{UTKFace,debias_domain_discriminative_RFW_MI_AUC_TAR_FAR, sex_race_PPB}. Second, a group of methods that attempt to explicitly produce models that factor in protected attribute information to close the performance gap between different demographics~\cite{debias_domain_discriminative_RFW_MI_AUC_TAR_FAR,debias_domain_discriminative_ROC_FAR,debias_domain_independent_training_BA_usage}.

Meanwhile, adopting a precise, universally applicable and acceptable metric for the degree of bias is both intrinsically difficult and important. Without loss of generality, three main approaches for bias assessment are based on either (1) classification accuracy across cohorts~\cite{debias_adversarial_uniform_confusion_ACC_unbalanced_dataset2, Dataset_bias,debias_domain_discriminative_RFW_MI_AUC_TAR_FAR}, (2) information leakage from protected attributes to prediction \emph{logits} of labels~\cite{DP, Image_caption_Bias_amplification_yz, debias_domain_independent_training_BA_usage}, or (3) estimated correlations  between  prediction \emph{logits} and protected attributes using shallow classifiers~\cite{debias_resampling_representation_bias_colored_MNIST, CAI_rz,UAI_rz}. Classification accuracy-based bias assessment metrics may not be accurate because accuracy across cohorts must be listed and compared together. Moreover, information leakage-based methods, such as \emph{Demographic Parity}, \emph{Equality of Odds}, and \emph{Equality of Opportunity}~\cite{DP} strictly define  fairness of a classification model as the independence between protected attributes and prediction \emph{logits} of labels, which may not be appropriate to compare debiased models directly, as discussed in~\cref{subsec:information_leakage}. Besides, quantitative metrics, relying on estimated correlations~\cite{arbitrary_correlation} using a shallow predictor, may find correlations even in unbiased data and would then mistakenly identify them as biased, as discussed in~\cref{subsec:correlation_construction}. Further, most of proposed bias assessment metrics are rarely used to evaluate other debiased models, as a mean to show universality. To the best of our knowledge, there is no universal way to assess the degree of bias for existing debiased models as these models have been evaluated on different benchmarks under varying conditions.

Prior work~\cite{debias_adversarial_uniform_confusion_ACC_unbalanced_dataset2,debias_domain_discriminative_ROC_FAR,debias_domain_discriminative_RFW_MI_AUC_TAR_FAR} mainly focus on bias mitigation instead of bias assessment, particularly neglecting representation/embedding level bias. Therefore, we propose an effective \emph{representation-level} method to assess the demographic bias of any pretrained backbone (possibly debiased) facial recognition systems and design a series of experiments as a protocol to validate the rationality and effectiveness of bias assessment metrics in face recognition. We use entropy to assess dataset bias and mutual information to assess model bias from learned representation extracted by backbone models, rather than simply establishing correlations by training a shallow predictor using \emph{logits}, as illustrated in~\cref{fig:idea}. We combine dataset bias and \emph{representation-level} model bias to comprehensively assess the percentage of remaining bias after a debiased backbone model, given the overall dataset bias. In other words, large remaining bias represents inferior debiasing performance. In this respect, our method can also help assess the bias using representations/embeddings in a layer-by-layer fashion inside any model. The proposed metric is not intended to be \emph{combined} with other debiasing methods, but to actually independently evaluate debiasing methods as shown in~\cref{Comparison_debiasing_models}. The effectiveness of our method is verified with experiments on Colored MNIST~\cite{MNIST}, FairFace~\cite{FairFace}, CelebA~\cite{CelebA} and synthetic datasets generated by StyleGAN2~\cite{styleGan2}. Our key contributions can be summarized as follows:

\begin{itemize}
    \item Theoretical and empirical arguments that bias assessment should be applied at the representation level instead of the \emph{logits} level.
    \item An \emph{independent} bias assessment metric at the representation level to help study bias mitigation. 
    \item A performance evaluation for a wide range of debiasing techniques using the proposed \emph{independent} metric.
    \item A synthetic dataset that mitigates the issue of insufficient samples in certain subsets.
    \item A categorization of different bias metrics.
\end{itemize}

\section{Related work}
\label{sec:related_work}
\noindent
\textbf{Protected attributes.}
\textit{Protected attributes are  qualities, traits or characteristics that, by law, cannot be discriminated against}~\cite{protected_attributes1}.
%Bias issues involving protected attributes referred to as demographic bias existing in deep learning are also very concerning.
%As shown in~\cite{Image_caption_Bias_amplification_yz}, algorithms will be more likely to generate a female caption for the person who is cooking or shopping. 
Many studies~\cite{Bias_report1, Bias_report2_race, FVRT} show that facial recognition systems have divergent recognition accuracy for different demographic groups. Meanwhile, using existing datasets~\cite{CelebA, IJBC}, which are dominated by sufficient samples in specific racial or sex groups, may also lead to unfairness against specific such groups.

\noindent
\textbf{Debiasing face recognition.} 
The fairness of face recognition may be dramatically impacted by the bias issues in existing datasets in terms of the \emph{long-tail distribution}~\cite{Long_tail} of demographic groups. To address bias issues, some studies introduce fairness into face recognition to mitigate demographic bias. The mainstream debiasing models belong to either (1) strategic sampling method~\cite{debias_resampling_representation_bias_colored_MNIST} via oversampling or re-weighting to keep the training data balanced across cohorts,  (2) representation disentanglement methods~\cite{debias_adversarial_Age_sex_race_biasness, debias_adversarial_uniform_confusion_ACC_unbalanced_dataset2, debias_adversarial_forgetting_yz} to remove the specific demographic attribute by \textit{adversarial training}, (3) domain adaptation methods~\cite{debias_domain_discriminative_RFW_MI_AUC_TAR_FAR, debias_domain_discriminative_ROC_FAR} for learning demographic-group invariant representations by maximizing the recognition performance of identity and minimizing the capability to predict  protected attributes using a discrimination loss, or (4) domain independent training method~\cite{debias_domain_independent_training_BA_usage} by learning an ensemble that constitutes separate classifiers per demographic group with \textit{representation sharing}.

\noindent
\textbf{Bias assessment metrics.} 
While many debiased models have been developed for face recognition, there has been limited progress in establishing an objective, quantitative and universally acceptable bias assessment metric. In particular, most of bias assessments rely on cross-cohort-terms based on \emph{classification accuracy}~\cite{debias_adversarial_uniform_confusion_ACC_unbalanced_dataset2}, \emph{False Positive Rate (FPR)}~\cite{debias_adversarial_projection_Bias_definition_FPR_DP}, \emph{Receiver Operating Characteristic (ROC)}~\cite{debias_domain_discriminative_ROC_FAR} and \emph{Area under the ROC Curve (AUC)}~\cite{debias_domain_discriminative_RFW_MI_AUC_TAR_FAR}. Besides, information leakage from protected attributes to predicted labels has also been used to assess bias. For example, \emph{Demographic Parity}, \emph{Equality of Odds}, and \emph{Equality of Opportunity}~\cite{DP} use independence between protected attributes and prediction \emph{logits} of labels to define fairness. Likewise, the difference between \emph{dataset leakage} (\ie the predictability of sex from ground truth labels) and \emph{model leakage} (\ie the predictability of sex from model predictions)~\cite{debias6_dataset_leakage_yz} have also been used to assess the bias. Similarly, \emph{bias amplification}~\cite{Image_caption_Bias_amplification_yz} is defined as the difference of \emph{bias score} (\ie the percentage of occurrences of a given outcome and a demographic variable in the corpus) between training data and testing data. Moreover, several bias assessment metrics based on the estimated correlation using a shallow network at the \emph{logits} level have been proposed. For example, \emph{dataset bias}~\cite{debias_resampling_representation_bias_colored_MNIST,representation_bias_RESOUND} captures the bias of a dataset, measured by the classification performance with the cross entropy loss. More generally, the metric used in~\cite{CAI_rz,UAI_rz} assesses bias based on prediction \emph{logits} to predict protected attributes from representations. However, all aforementioned metrics are implicitly or explicitly based on accuracy or \emph{logits} loss after the predictor at the \emph{logits} level instead of the representation/embedding level, and therefore we call them \emph{logits}-level bias assessment metrics. 

\emph{Distance correlation $dcor^2$}~\cite{correlation_distance} has been used to assess the bias at the representation level in~\cite{BR_net_correlation_distance_bias_metric} . However, in~\cite{BR_net_correlation_distance_bias_metric}, the usage of \emph{distance correlation} only considers model bias from the representations without a correction of dataset bias if the distribution of protected attributes varies as discussed in~\cref{colored_MNIST}, and yields more variation than our proposed metric as discussed in \cref{Comparison_debiasing_models}.

% \red{Furthermore, since the usage and names of different bias assessment metrics are intricate without a standard benchmark to unify and evaluate different debiased models together, we show that an accurate \emph{independent} bias assessment metric can assess demographic bias from learned representation and protected attribute labels directly, and can be used in a number of debiased model evaluation tasks.}

\section{Representation-Level Bias Assessment}
As discussed in~\cref{sec:related_work}, most of the existing bias assessment methods depend on training shallow predictors that could overfit to \emph{spurious} correlations between the input data and protected attributes, as illustrated in~\cref{fig:idea}. Our goal is therefore to develop an \emph{independent} (\ie method-agnostic) bias assessment metric which can be applied at the representation/embedding level of pretrained (possibly debiased) models, and considers training dataset bias.

Consider a face recognition task for which, given a dataset $\mathcal{D}$ containing instances $(x_i, y_i, z_i)$, where $x_i \in \mathcal{X}$ is an image annotated with a set of task-specific labels $y_i \in \mathcal{Y}$ (e.g. identity), and other protected attributes $z_i \in \mathcal{Z}$ (e.g. sex\footnote{In this paper, due to the available annotations, we assume that sex is binary, but the work can be extended to non-binary sex annotations.}), the representation learning network $F_\theta: \mathcal{X} \to \mathcal{R}$ parametrized by $\theta \in \Theta$ first produces a learned representation $r_i \in \mathcal{R}$, and then a classifier $C_\phi: \mathcal{R} \to \mathcal{Y'}$ with parameters $\phi \in \Phi$ produces the predicted label $y'_i \in \mathcal{Y'}$. The learned representation $r_i \in \mathcal{R}$ produced by representation learning sub-network $F_\theta$ may contain information about $z_i$, due to the end-to-end nature of training, which encourages models to exploit any information (including protected attributes) if it leads to lower empirical loss. \emph{Demographic parity} (DP)~\cite{DP} seeks to find information leakage between $\mathcal{Z}$ and $\mathcal{Y'}$.

\noindent
\textbf{Definition 1:} DEMOGRAPHIC PARITY. A classification model $\hat{T}$ is said to satisfy \emph{demographic parity} if predicted label $Y' = \hat{T}(X)$ and protected attribute $Z$ are independent.

However, DP does not completely ensure fairness~\cite{debias1_no_bam_DP} since the \emph{logits}-level parity can arise naturally when there is little training data for one protected attribute $z_i$, and may impair the achievable utility of better classification accuracy since some correct predictions may contradict DP in general if the testing dataset is not strictly balanced, which will be further elaborated in~\cref{subsec:information_leakage}. In order to overcome these drawbacks, we therefore propose:

\noindent
\textbf{Definition 2:} REPRESENTATION-LEVEL DEMOGRAPHIC PARITY. A classifier $\hat{T}$ is said to satisfy \emph{representation-level demographic parity} if learned representation $\hat{R} = \hat{F}(X)$ and protected attribute $Z$ are independent.

Representation-level demographic parity means that for all values of the protected attributes $Z$: $ P(\hat{F}(X) = \hat{r}) = P(\hat{F}(X) = \hat{r} | Z = z) $ where $\hat{F}$ is a representation learning sub-network. Mutual information (MI) which is widely used in representation disentanglement and debiasing~\cite{debias_domain_discriminative_RFW_MI_AUC_TAR_FAR, debias4_extreme_bias_MI_colored_MNIST, mutual_information_mitigation}, is then a natural approach for assessing the mutual dependence between $\mathcal{R}$ and $\mathcal{Z}$, and produce an information-theoretic fairness score. Independence is achieved when the representation space $\mathcal{R}$ contains no information about protected attributes $\mathcal{Z}$.
%On the other extreme, when there is a deterministic function $M$ between $\mathcal{R}$ and $\mathcal{Z}$, \ie $\mathcal{Z} = M(\mathcal{R})$, $\mathcal{R}$ conveys all information of $\mathcal{Z}$, which is extremely biased. 
More generally, we can say that $B \propto I(R, Z)$, where $B$ is the representation-level bias and $I$ is mutual information.

Facial recognition bias~\cite{survey} stems from a biased trained model $T_{biased}$ and/or an imbalanced training dataset $\mathcal{D}_{biased}$. Representation-level bias reveals the degree of bias for the model $T_{biased}$ reflected in the learned representation $R$ extracted by the feature extraction sub-network $F$ inside $T_{biased}$. We use mutual information between $R$ and $Z$ to estimate the representation-level bias for the biased trained model $T_{biased}$, \ie $I(R, Z)$. Furthermore, since a more imbalanced training dataset leads to more bias in the trained model, we use the entropy of $Z$ to assess the imbalance of the dataset $\mathcal{D}_{biased}$, \ie $H(Z)$. Greater entropy implies a more balanced dataset. Therefore, we define the representation-level bias as follows.

\noindent
\textbf{Definition 3:} REPRESENTATION-LEVEL BIAS (RLB). The representation-level bias $B$ of a classification model $T$ trained with a dataset $\mathcal{D}$, with respect to protected attribute $Z$, is defined as,
\begin{align}
    B(R, Z) = \frac{I(R, Z)}{H(Z)},
\label{representation_bias}
\end{align}
\hspace*{\fill} \\
where $H(Z)$ is the entropy of $Z$, estimated empirically by:
\begin{align}
    H(Z) = -\frac{1}{|\mathcal{Z}|}\sum_{z\in \mathcal{Z}}^{}\log_{}{P_z},
\end{align}
and $I(R,Z)$ is the mutual information between $R$ and $Z$, estimated based on Definition 3.1 in ~\cite{MINE}:
\begin{align}
\label{eq_lower_bound_estimation}
    I(R,Z) = \sup_{\theta\in\Theta} \mathop{{}\mathbb{E}_{\mathbb{P}_{   \mathcal{RZ}}}} [T_\theta] - \log(\mathop{{}\mathbb{E}_{{\mathbb{P}_{\mathcal{R}} \otimes {\mathbb{P}_{\mathcal{Z}}}}}} [e^{T_\theta}]),
\end{align}
which estimates MI by training a neural network $T_\theta$ to distinguish between joint samples $\mathbb{P}_{\mathcal{RZ}}$ and the product of marginals $\mathbb{P}_{\mathcal{R}} \otimes \mathbb{P}_{\mathcal{Z}}$, of random variables $R$ and $Z$. The ratio $B$ is bounded ($[0,1]$) and easy to interpret, rather than the uncertain and negative range for mutual information minus entropy (since entropy is greater than mutual information). Mutual information neural estimation (MINE)~\cite{MINE} offers a lower-bound based on the \emph{Donsker-Varadhan representation}~\cite{MINE_base} of KL-divergence. 

We improve the mutual information estimate of ~\cite{MINE} by adding a mapping network followed by the statistics network to improve the robustness. Given a pair of $(r,z) \in (\mathcal{R},\mathcal{Z})$, the non-linear mapping networks $f_{\lambda_1} : \mathcal{R} \to \mathcal{W}$ and $f_{\lambda_2} : \mathcal{Z} \to \mathcal{S} $ first produce $w \in \mathcal{W}$ and $s \in \mathcal{S}$. The mapping networks $f_{\lambda_1}$ and $f_{\lambda_2}$ are implemented using one fully connected layer with parameters $\lambda_1$ and $\lambda_2$. Input dimensionality are adapted with $(\mathcal{R},\mathcal{Z})$ and output dimensionality is kept same. Then, $k$ minibatch samples are draw from the joint distribution, \ie
$(w^J_1,s^J_1),...,(w^J_k,s^J_k) \sim \mathbb{P}_{\mathcal{WS}}$. Similarly, we keep the same $k$ samples from the marginal distribution $\mathbb{P}_{\mathcal{S}}$, \ie $s^J_1,..,s^J_k \sim \mathbb{P}_{\mathcal{S}}$, and draw $k$ samples from the marginal distribution $\mathbb{P}_{\mathcal{W}}$, \ie
$w^M_1,..,w^M_k \sim \mathbb{P}_{\mathcal{W}}.$
The statistics network $g_{\theta} : \mathcal{W} \times \mathcal{Z} \to \mathbb{R}$ parametrized by  $\theta \in \Theta$ is designed to evaluate the lower-bound of mutual information as a real number. Mutual information approximation is estimated by,
\begin{align}
    I_{\theta}(W, S) = \frac{1}{k}\sum_{i=1}^{k} o^J_i - \log(\frac{1}{k}\sum_{i=1}^{k} e ^ {o^M_i}),
\end{align}
where
\begin{align}
     o^J_i = g_{\theta} (w^J_i, s^J_i), \quad\textrm{and}\quad  o^M_i = g_{\theta}  (w^M_i, s^J_i). 
\end{align}
In each training iteration, the gradient propagates through the statistics and the mapping networks, and the parameters $[\lambda_1, \lambda_2, \theta]$ are updated by the gradient of loss function 
\begin{align}
\begin{split}
        L = -(\overline O^J_i - \frac{\overline O^M_i }{EMA^M_t}),
\end{split}
\end{align}
where $EMA^M_t$ is the exponential moving average (EMA) of marginal sample outputs, \ie
\begin{align}
\begin{split}
        EMA^M_t=\left\{
    \begin{array}{lr}
    \overline O^M_t, & t=0 \\
    \alpha \overline O^M_t + (1 - \alpha)\overline O^M_{t-1}, & t>1  \\
    \end{array}
    \right.
\end{split}
\end{align}
where $\overline O^M_t$ is the moving average at iteration $t$ and $\alpha$ is smoothing coefficient. We optimize by simultaneously estimating and maximizing the mutual information until convergence as follows:
\begin{align}
    I(R,Z) = \max_{\theta \in \Theta} I_{\theta}(W, S).    
\end{align}
Finally, Representation-Level Bias (RLB) $B(R,Z)$ can be given according to \textbf{Def. 3}.

% The overview for the model is shown in~\cref{fig:model}.

\section{Comparing Bias Assessment Metrics}
\label{sec:logits-level bias}

\begin{table}[]
% \footnotesize
\scriptsize
% \tiny
\centering
\caption{Taxonomy of different bias metrics.}
\begin{tabular}{lll}
\hline
Taxonomy                 & Usage & Examples                                                                                                             \\ \hline
Accuracy across cohorts  & $R, Y$  & \makecell[l]{Standard deviation of accuracy~\cite{DB-VAE, GAC}, \\ other metrics across cohorts\\ (ROC~\cite{debias_domain_discriminative_ROC_FAR}, AUC~\cite{debias_domain_discriminative_RFW_MI_AUC_TAR_FAR}, F1 score~\cite{BR_net_correlation_distance_bias_metric}).} 
\\ \hline
Information leakage      & $Z, Y'$ & \makecell[l]{Demographic Parity, Equality of Odds, \\ Equality of Opportunity~\cite{DP}, \\ Dataset leakage, Model leakage~\cite{debias6_dataset_leakage_yz}, \\ \emph{Bias amplification}~\cite{Image_caption_Bias_amplification_yz}.}                                    \\ \hline
Estimated correlation    & $R, Z$  & \makecell[l]{Dataset bias~\cite{debias_resampling_representation_bias_colored_MNIST}, \\ \emph{logits}-level loss~\cite{UAI_rz, CAI_rz}.}                                                                                      \\ \hline
Statistical dependence & $R, Z$  & \makecell[l]{\emph{Distance Correlation}~\cite{correlation_distance}, \\ \textbf{Representation-Level Bias (RLB)}.}                                                                                     \\ \hline
\end{tabular}
\label{fig:metrics}
\vspace{-.7cm}
\end{table}

Having introduced RLB, we will theoretically discuss the advantages of RLB compared to several representative bias assessment metrics at the \emph{logits} level. \cref{fig:metrics} summarizes the comparison.

\subsection{Information Leakage Fairness Criterion}
\label{subsec:information_leakage}
Information leakage-based metrics assess bias by measuring information leakage from $\mathcal{Z}$ to $\mathcal{Y'}$. For example, \emph{Demographic Parity}, \emph{Equality of Odds}, and \emph{Equality of Opportunity}~\cite{DP} use independence between protected attributes $\mathcal{Z}$ and prediction \emph{logits} of labels $\mathcal{Y'}$, which are commonly used as the strict definition of fairness for data-driven classification models. \emph{Demographic parity} (DP) requires the classification to be independent of protected attributes. Specifically, besides the predictor $P$ estimating $Y$ as accurately as possible, an additional adversarial network $G$ is introduced to predict a value for $Z$ from $Y'$. DP is achieved when limiting any information about $Z$ leaking to $Y'$. However, as argued in~\cite{debias1_no_bam_DP}, DP has two limitations. First, the fairness may not be completely ensured under DP since the \emph{logits}-level parity can arise naturally with little training data of $z_i$. By contrast, representation-level DP does not arise naturally since it is an ideal criterion which requires independence of high-dimensional learned representations and protected attributes, which is naturally unattainable in practice. Further, in contrast to $Z$, $R$ are high dimensional vectors with much higher capacity to tolerate noise than predicted labels $Y'$ in original demographic parity. Second, DP may harshly forbid some correct predictions if they violate the criterion in general, which hinders achievable better classification accuracy. The failure case of pursuing DP is that some correct predictions may be forced to be incorrect since DP requires strict probability equality across cohorts. Further, compared to the harsh DP, RLB is a soft metric.

% \begin{figure*}[t!]
% \begin{center}
% %\fbox{\rule{0pt}{2in} 
% %\rule{0.9\linewidth}{0pt}}
%   \includegraphics[width=0.9\linewidth]{sections/figure/pipeline.png}
% \end{center}
%   \caption{An overview of the proposed method. The presented method consists of two mapping networks and one statistics network, indicated in yellow. Given learned representation $R$ and protected attributes $Z$, the mapping networks $f_{\lambda_1} : \mathcal{R} \to \mathcal{W}$ and $f_{\lambda_2} : \mathcal{Z} \to \mathcal{S} $ produce $w \in \mathcal{W}$ and $s \in \mathcal{S}$. The corresponding representation maps are then sampled for joint samples and marginal samples. Finally, representation-level bias is estimated by statistics network $g_\theta$ with mutual information.}
% \label{fig:model}
% \end{figure*}

\subsection{Correlation Estimation}
\label{subsec:correlation_construction}
Correlation estimation-based metrics assess bias by estimating the correlation between $\mathcal{R}$ and $\mathcal{Z}$ using a shallow network.
As mentioned in~\cite{CAI_rz,UAI_rz} to assess  model bias by \emph{logits}-level loss, they try to estimate the correlation by training a mapping $M$ from the family of shallow predictors  $P_{\psi}$ parametrized by  $\psi \in \Psi$ such that $Z^l \approx M(R^k)$, where $M \in \{P_\psi\}_{\psi \in \Psi}$, $R^{k}$ is the learned representation with dimension $k$ and $Z^{l}$ is the protected attributes with dimension $l$ $(l<k)$. Model bias can then be assessed at the \emph{logits} level by minimizing the loss or maximizing prediction accuracy from the predictor $P_{\hat{\psi}}$ that learns to predict $\mathcal{Z}$ from $\mathcal{R}$, with  parameters $\hat{\psi}$ such that 
\begin{align}
    \hat{\psi} = \underset{\psi \in \Psi}{\operatorname{arg\,min}}~Loss( P_{\psi}(R^{k}), Z^{l}).    
\end{align}
However, this correlation is unstable and capricious since, the predictor $P_{\hat{\psi}}$ with  parameters $\hat{\psi}$ is easily trained as a projection from $\mathcal{R}^{k}$ to $\mathcal{Z}^{l}$, \ie from a high dimensional space to a low dimensional space.

Furthermore, we may arbitrarily construct a spurious or uncorrelated representation space $\Tilde{\mathcal{R}}^k$ with the same dimension $k$ as real representation space $\mathcal{R}^k$ to confuse the shallow network $P_{\psi}$, as shown in~\cref{robustness}. The confused shallow network may also find a correlation between $\Tilde{\mathcal{R}}^k$ and $\mathcal{Z}^l$ such that $Z_i^l \approx \Tilde{M}(\Tilde{R}^{k}_{i})$, where $\Tilde{M} \in \{P_\psi\}_{\psi \in \Psi}$. Unfortunately, the spurious mapping $\Tilde{M}$ would offer a minimum loss or a maximum prediction accuracy as the degree of bias even for the uncorrelated representation and protected attributes.

\begin{figure*}[htbp]
  \centering
%     \subfloat[][Random digit examples, gently biased (\textbf{top}) and extremely biased (\textbf{bottom}).]{\label{fig:colored_sample}\includegraphics[width=.8\linewidth]{sections/figure/color_MNIST_sample.png}}
% \\
\begin{minipage}[b]{.3\linewidth}
    \centering
    \subfloat[][Convergence lines of mutual information estimation.]{\label{fig:colored_converge}\includegraphics[width=5cm]{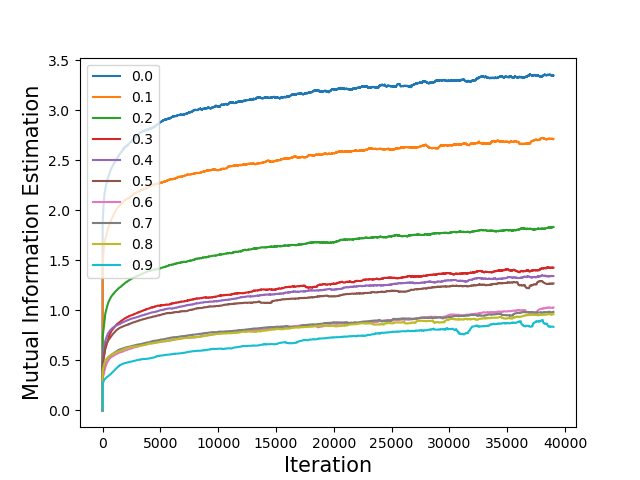}}
\end{minipage}
\begin{minipage}[b]{.3\linewidth}
    \centering
    \subfloat[][Bias assessment of testing dataset with different standard deviations.]{\label{fig:dataset_bias}\includegraphics[width=0.9\linewidth]{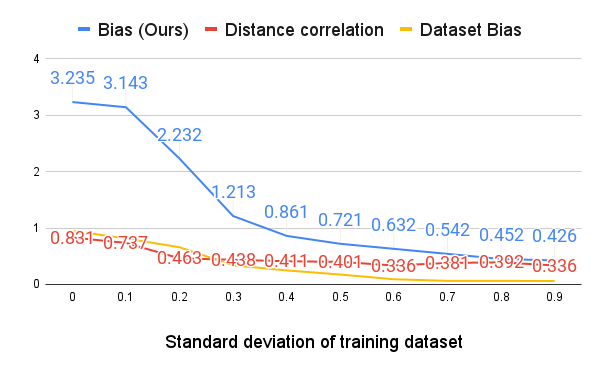}}
\end{minipage}
\begin{minipage}[b]{.3\linewidth}    
    \subfloat[][Bias assessment of testing dataset with the fixed standard deviation.]{\label{fig:representation bias}\includegraphics[width=0.9\linewidth]{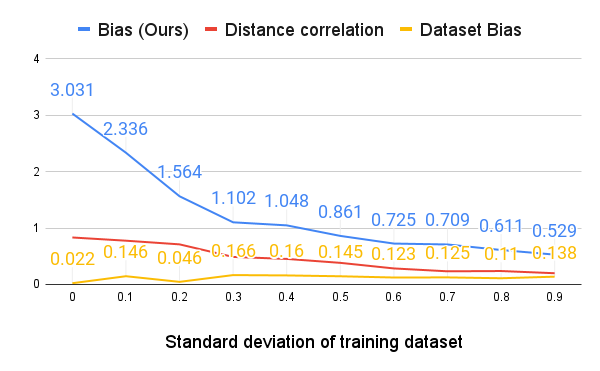}}
\end{minipage}
   \caption{Verification experiments on Colored MNIST Dataset.}
\label{fig:short}
\vspace{-.7cm}
\end{figure*}

By contrast, the principal idea of RLB is that the correlation between $\mathcal{R}$ and $\mathcal{Z}$ should be independently estimated by mutual information, instead of a neural network in~\cite{CAI_rz,UAI_rz}. The drawback that an auxiliary neural network in \logits-level loss may produce spurious correlations is addressed by mutual information lower-bound estimation, for which consistency of the \emph{Donsker-Varadhan representation} and the network parameters choice over $\{\theta\in\Theta\}$ for MI supremum in~\eqref{eq_lower_bound_estimation} are proved in~\cite{MINE}. The limitation of \emph{logits}-level loss using a neural network predictor is a preconceived latent assumption that the correlation is determined by a mapping $M$ as a neural network with a specific architecture and   parameters, such that $Z^l \approx M(R^k)$ where $M = \{P_\psi\}_{\psi \in \Psi}$. By contrast, using mutual information, we do not care about what the mapping $M$ is explicitly so that we can relax the mapping $M$ of $\mathcal{Z}$ from $\mathcal{R}$ without regarding it as a specific neural network. Furthermore, the neural network in RLB initializes with the lowest estimated bias, and in each iteration, strives to increase it by exhaustively traversing different mapping functions $T_\theta$, as shown in \cref{fig:colored_converge}. The curve eventually converges to the greatest estimated bias which is approximately the lower bound of the actual bias, which explains the fact that, in~\cref{tab:synthesized_representation2}, the estimated bias stays at the lowest point with synthesized representations. Contrastively, there is no lower-bound guarantee for the \logits-level metrics based on the prediction accuracy or \logits~loss so that it may exceed the actual bias.

\begin{figure}[htbp]
\centering
\includegraphics[width=0.8\linewidth]{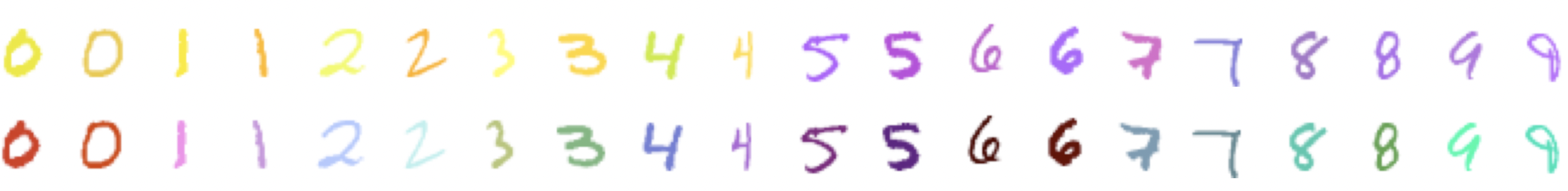}
 \caption{Examples of Colored MNIST, moderately biased (\textbf{top}) and extremely biased (\textbf{bottom}).}
  \label{fig:colored_sample}
  \vspace{-.7cm}
\end{figure}

% \begin{figure}[htbp]
% \centering
% \includegraphics[width=0.8\linewidth]{sections/figure/color_MNIST_sample.png}
%  \caption{Random digit examples, gently biased (\textbf{top}) and extremely biased (\textbf{bottom}).}
%   \label{fig:black}
% \end{figure}

% \begin{figure}[htbp]
% \centering
% \includegraphics[width=0.8\linewidth]{sections/figure/colored_mnist_convergence.png}
%  \caption{Representation-level bias for sex and race of FairFace dataset.}
%   \label{fig:black}
% \end{figure}

% \begin{figure}[htbp]
% \centering
% \includegraphics[width=0.8\linewidth]{sections/figure/dataset_bias_Colored_MNIST.png}
%  \caption{Representation-level bias for sex and race of FairFace dataset.}
%   \label{fig:black}
% \end{figure}

% \begin{figure}[htbp]
% \centering
% \includegraphics[width=0.8\linewidth]{sections/figure/representation_bias_Colored_MNIST.png}
%  \caption{Representation-level bias for sex and race of FairFace dataset.}
%   \label{fig:black}
% \end{figure}

\section{Experimental Evaluation}
\label{sec:experiments}

Prior work~\cite{UTKFace, debias_domain_discriminative_RFW_MI_AUC_TAR_FAR, sex_race_PPB} argues that any bias assessment metrics must be capable of evaluating both dataset (im)balance and model bias. Therefore, we empirically demonstrate that our \emph{independent} bias assessment metric at the representation level is more effective than other bias assessment metrics at the \emph{logits} level in these two aspects with experiments on (1) Colored MNIST~\cite{MNIST}, (2)  FairFace~\cite{FairFace}, (3) CelebA~\cite{CelebA}, and (4) synthetic datasets generated by StyleGAN2~\cite{styleGan2}. First, a number of synthesized representations are used to test the robustness of our method compared to the estimated-correlation metrics, \ie \logits-level loss~\cite{CAI_rz,UAI_rz} in \cref{robustness}. Second, we verify that our method outperforms another estimated-correlation metric, \ie \emph{dataset bias}~\cite{representation_bias_RESOUND, debias_resampling_representation_bias_colored_MNIST} and a statistical-independence metric, \ie \emph{distance correlation} ${dcor}^2$~\cite{correlation_distance} by showing different bias scores for imbalanced datasets in~\cref{colored_MNIST} and reflecting the model bias at the representation level,  which is further demonstrated in~\cref{subsec:fairface} and~\cref{synthetic_dataset}. Finally, we show that RLB is generic and capable of evaluating different debiased models compared to the methods based on accuracy across cohorts, information leakage, \ie \emph{bias amplification} (BA)~\cite{Image_caption_Bias_amplification_yz} and statistical dependence, \ie ${dcor}^2$~\cite{correlation_distance} in~\cref{Comparison_debiasing_models}.

% several representative bias assessment metrics for each category, \emph{dataset bias}~\cite{debias_resampling_representation_bias_colored_MNIST,representation_bias_RESOUND}, \logits-level loss~\cite{CAI_rz,UAI_rz} (estimated correlation), \emph{bias amplification}~\cite{Image_caption_Bias_amplification_yz} (information leakage), \emph{distance correlation}~\cite{BR_net_correlation_distance_bias_metric} (statistical dependence) and accuracy across cohorts,

\subsection{Robustness Against Spurious Correlations}
\label{robustness}
To verify the robustness of RLB and show that correlations could be found between both true or spurious protected attributes and both true or spurious representations, we introduce several synthesized representations using the Aligned \& Cropped subset of CelebA dataset~\cite{CelebA} since the learned representations extracted from cropped images focus more on demographic appearances (\eg hair type, colors) and avoid the interference from other features (clothes). First, we train a ResNet-50~\cite{ResNet50} to recognize attributes on CelebA dataset and use the qualified network $F$ to construct learned representation space $\mathcal{R}$. Then, we introduce several synthesized representations, including (1) $R_S$, shuffling learned representation over the feature dimension; (2) $R_G$, generating unpaired representations from a different but same-sample-size batch; (3) $Z_S$, shuffling protected attributes over the samples; and (4) $Z_G$, generating unpaired but overall-entropy-unchanged protected attributes labels, to confuse the bias assessment metric based on \emph{logits}-level loss.

In~\cref{tab:synthesized_representation2}, we compare the proposed RLB with the \emph{logits}-level loss (estimated-correlation metric) for sex bias. The results show that the correlations exist whether or not the synthesized representations are applied. There is no discrimination capability for testing prediction accuracy since several accuracy with synthesized representations approximate or exceed the accuracy without synthesized representations. Furthermore, the \emph{logits} loss with synthesized representations is expected to be at least greater than that without synthesized representations since the spurious representation space $\Tilde{\mathcal{R}}^k$ is fabricated at random and should be uncorrelated to protected attributes space $\mathcal{Z}^l$. However, the \emph{logits}-level loss is crude to construct spurious correlations between unrelated samples. On the other hand, as we add spurious correlations, RLB declines, which means the correlation constructed by mutual information only exists between $\mathcal{R}$ and $\mathcal{Z}$ instead of any spurious representation space.

\begin{table}[htb]
\centering
% \footnotesize
\scriptsize
% \tiny
\caption{Comparison between correlations established by predictor and mutual information.}
\begin{tabular}{|c|c|c|c|c|c|}
\hline
\multicolumn{1}{|l|}{\multirow{2}{*}{}} & \multicolumn{1}{c|}{\textbf{Normal}}      & \multicolumn{1}{c|}{\multirow{2}{*}{\bm{$R_S$}}} & \multicolumn{1}{c|}{\multirow{2}{*}{\bm{$R_G$}}} & \multicolumn{1}{c|}{\multirow{2}{*}{\bm{$Z_S$}}} & \multicolumn{1}{c|}{\multirow{2}{*}{\bm{$Z_G$}}} \\
\multicolumn{1}{|l|}{}                  & \multicolumn{1}{c|}{\textbf{Correlation}} & \multicolumn{1}{c|}{}                   & \multicolumn{1}{c|}{}                   & \multicolumn{1}{c|}{}                   & \multicolumn{1}{c|}{}                   \\ \hline
\makecell[c]{Testing Acc.} & 99.7 & 63.8 & 99.9 & 92.3 & 99.8 \\ \hline
\emph{Logits} Loss & 0.054     & 0.412     & 0.23e-05 & 0.179  & 0.026  \\ \hline
\textbf{Bias (Ours)} & \textbf{0.874}          & \textbf{2.42e-05}     & \textbf{0.34e-05} & \textbf{1.28e-05}  & \textbf{1.09e-05}  \\ \hline
\end{tabular}
\label{tab:synthesized_representation2}
\vspace{-.5cm}
\end{table}

\subsection{Colored MNIST}
\label{colored_MNIST}
To obtain intuitive insights and demonstrate the effectiveness of RLB for evaluating the (im)balance of training datasets and capturing dataset bias, we conduct experiments on a modified version of MNIST~\cite{MNIST}, Colored MNIST, where assigned colors are sampled from digit-dependent distributions. Since the digits are tied up with the colors, color classification can facilitate digit classification; and therefore, Colored MNIST is biased with color representations and the intensity can be controlled by the color assignment scheme. The moderately biased case (assigning two colors to two groups of digits) and extremely biased case (assigning ten distinct colors to each digit) are shown in~\cref{fig:colored_sample}.

\noindent
\textbf{Experiment Setup.}
We introduce color bias by assigning RGB colors $z_i = (r_i, g_i, b_i)$ to each digit as the center color and provide a standard deviation (STD) $\sigma$ as its range; therefore, the color spectrum covered by each digit is $(r_i\pm\sigma, g_i\pm\sigma, b_i\pm\sigma)$. Increasing the STD $\sigma$ reduces bias since a larger STD will produce more overlap between the colors of different categories, thereby reducing the discriminability of colors. We train a LeNet-5 CNN~\cite{lenet} to recognize digits on Colored MNIST training set with different STDs $\sigma_{train}^{(i)}$ and use the qualified representation learning networks $F_{\sigma_{train}^{(i)}}$ to construct learned representation space $\mathcal{R}$ for representation-level bias assessment usage on the testing set with both different STDs $\sigma_{test}^{(i)}$ but same with training data and a fixed STD $\sigma_{test} = 0.5$. Experiments with different STDs in testing data demonstrate that our method reflects the degree of bias in dataset, and other experiments with a fixed STD verify that the estimation of mutual information reflects bias in the trained model since the color entropy of the testing data is same, \ie the denominator in \eqref{representation_bias} is same and the bias issues can only come from different biased models trained with different STDs.

\noindent
\textbf{Results.}
\cref{fig:colored_converge} shows that MI estimation, for a training dataset with different STDs and a testing dataset with a fixed STD, converge as several straight lines. \cref{fig:dataset_bias} shows that in the case that the STD of training and testing datasets are similar, both \emph{dataset bias}~\cite{debias_resampling_representation_bias_colored_MNIST} and RLB decrease as the STD increases. However, ${dcor}^2$ does not yield this trend in high STD since it only considers model bias without a correction of dataset bias. In \cref{fig:representation bias}, the decrease of \textit{dataset bias}~\cite{debias_resampling_representation_bias_colored_MNIST} with the increase of STD does not happen with a fixed STD $\sigma_{test}$, and in nature it only assesses the bias from testing dataset with $\sigma_{test} = 0.5$. On the other hand, the bias in the trained model is reflected in the representations of the testing dataset and captured by RLB as increase of the STD $\sigma_{train}$ of training dataset reduces RLB.

\begin{figure}[htbp]
\centering
\begin{minipage}[b]{.45\linewidth}
    \centering
    \subfloat[][{Sampled FairFace datasets.}]{\label{fig:black}\includegraphics[width=0.9\linewidth]{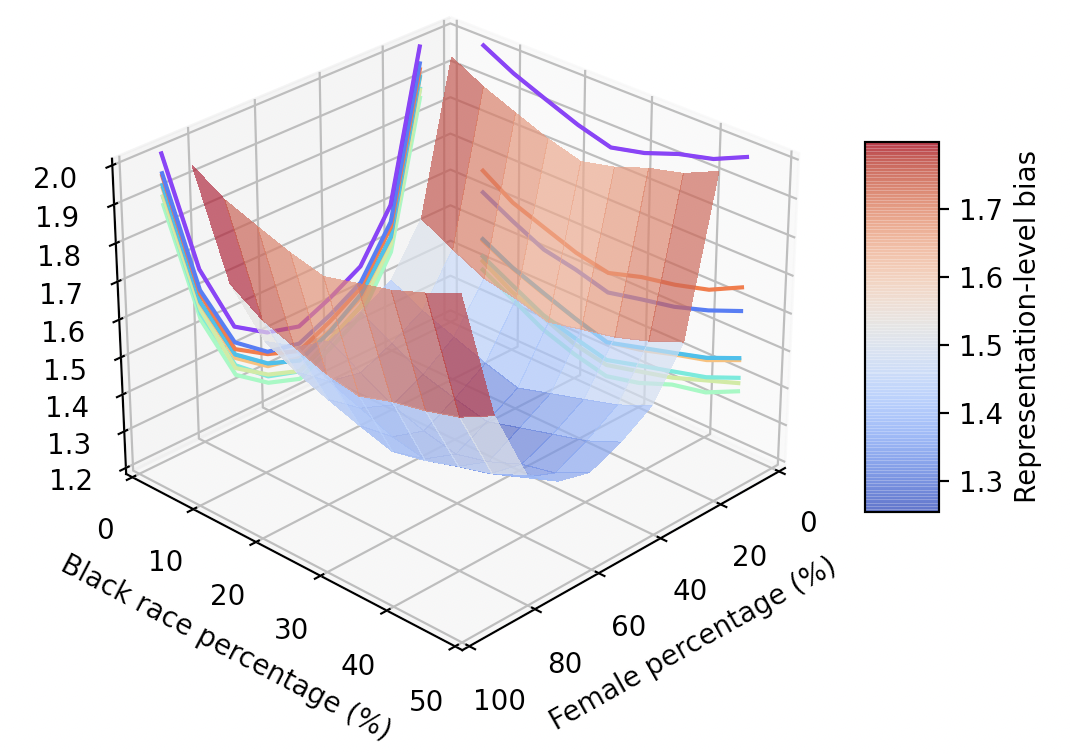}}
\end{minipage}
\begin{minipage}[b]{.45\linewidth}    
  \centering
    \subfloat[][Synthetic datasets.]{\label{fig:entropy}\includegraphics[width=.9\linewidth]{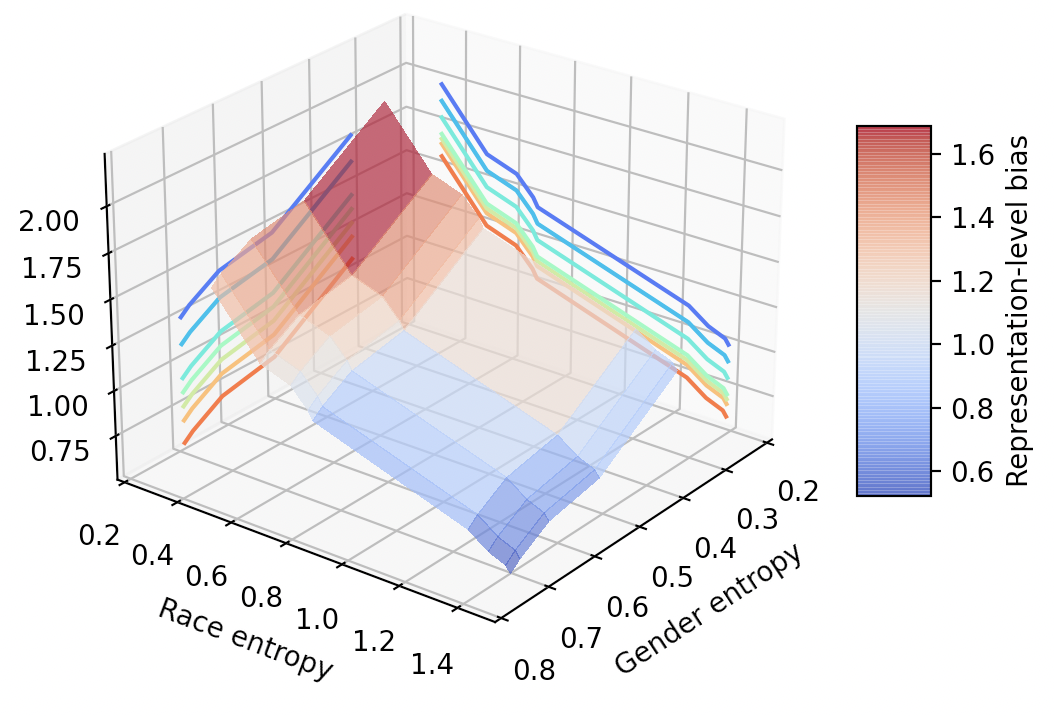}}
\end{minipage}
    \caption{Representation-level bias for sex and race.}
\label{fig:RLB_Sex_Race}
\vspace{-.5cm}
\end{figure}

\subsection{FairFace Dataset}
\label{subsec:fairface}
Sampled datasets from FairFace dataset~\cite{FairFace} are used to assess the bias induced by imbalanced datasets and explain the discrepancy due to imbalanced training~\cite{debias_adversarial_uniform_confusion_ACC_unbalanced_dataset2, survey}.

\noindent
\textbf{Experiment Setup. }
We emulate sex and racial bias by controlling female percentage $s^f_i$ in all sex attributes or black race percentage $r^b_i$ among all race attributes (Black, White, Indian and East Asian in this experiment). Approaching a balance point ($s^f_i = 0.5$ and $r^b_i = 0.25$) reduces the bias since more balanced datasets imply less bias. We train a ResNet-34~\cite{ResNet50} to recognize identities on sampled FairFace datasets with different $s^f_i$ and use the qualified representation learning networks $F_{s^f_i}$ to extract representation $\mathcal{R}$ for representation-level bias assessment usage. Further, a dataset considering imbalance of multiple protected attributes, which is similar to other imbalanced datasets~\cite{CelebA, IJBC} and the real world demographic distribution, is sampled based on $s^f_i$ and a specific race percentage $r_i$, such as black race percentage $r^b_i$. Meanwhile, the other races are sampled equally. %\red{To clarify, such imbalanced issue in multiple protected attributes is only guaranteed in the whole dataset level, without insurance of the desired race distribution inside sex due to the limited training samples after sampling. }

\begin{figure}[htbp]
\centering
\includegraphics[width=1.0\linewidth]{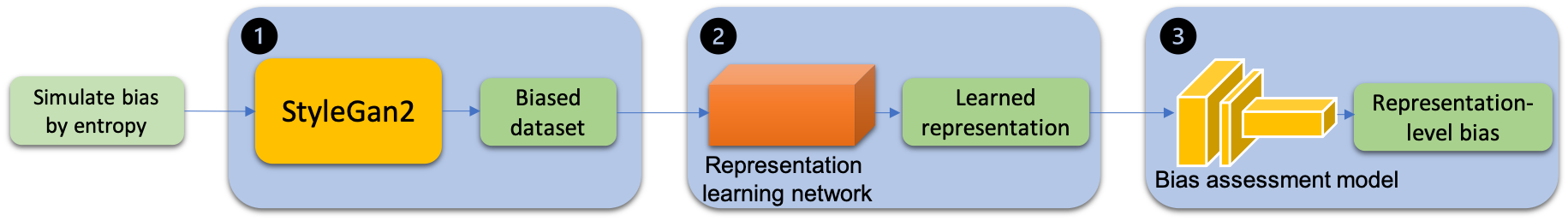}
 \caption{Three-stage experiment pipeline of Synthetic datasets.}
   \label{fig:stylegan2_whole_pipeline}
  \vspace{-.3cm}
   \end{figure}

\noindent
\textbf{Results. }
The results of RLB considering multiple protected attributes are shown in~\cref{fig:black}, as a \emph{basin} shape in 3-D space, with the lowest RLB when $s^f = s^m = 0.5$ and $r^b = r^w = r^a = r^i = 0.25$, which is desired since the dataset is balanced in both race and sex at these percentages. Furthermore, \cref{fig:black} shows that the projection curve of the \emph{basin} shape in the female percentage plane is a \emph{V} curve, but the projection curve in the black race percentage plane is flatter, which means that changes in female percentage has a stronger effect on RLB than changes in black race percentage. This difference is also desirable since female percentage of two sex groups may lead to greater imbalance than black race percentage of four race groups.

\begin{figure}[htbp]
\centering
\includegraphics[width=0.6\linewidth]{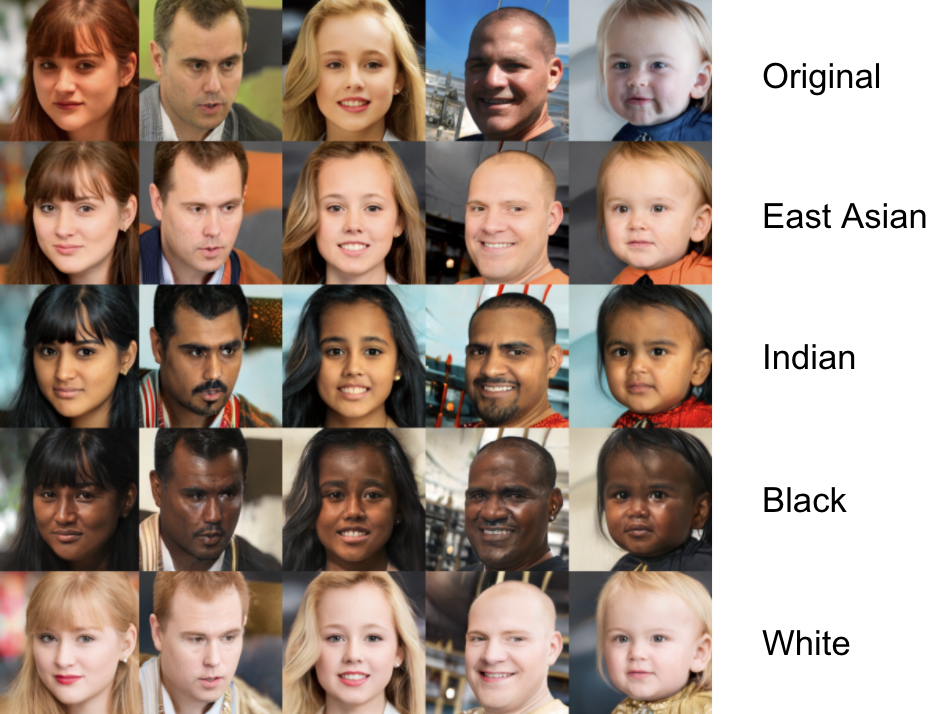}
 \caption{Examples of Synthetic datasets.}
   \label{fig:generated_dataset_sample}
   \vspace{-.6cm}
\end{figure}

\subsection{Synthetic Datasets Generated Using StyleGAN2}
\label{synthetic_dataset}
Due to the insufficiency of samples after splitting by several protected attributes, existing datasets may not be sufficient. Inspired by~\cite{multiple_style_direction}, we propose a new synthetic dataset generated by StyleGAN2~\cite{styleGan2} with a more complicated distribution, to facilitate experimental rather than observational analyses of the presented method.

% \begin{figure*}[htbp]
% \centering
% \begin{minipage}[b]{.32\linewidth}
%     \centering
%     \subfloat[][Bias (bias amplification~\cite{Image_caption_Bias_amplification_yz}).]{\label{subfig:BA}\includegraphics[width=5cm]{sections/figure/box_BA.png}}
% \end{minipage}
% \begin{minipage}[b]{.32\linewidth}
%     \centering
%     \subfloat[][on~\cite{correlation_distance}.]{\label{subfig:dcor}\includegraphics[width=0.9\linewidth]{sections/figure/box_dcor.png}}
% \end{minipage}
% \begin{minipage}[b]{.32\linewidth}    
%     \centering
%     \subfloat[][Bias (Ours).]{\label{subfig:ours}\includegraphics[width=0.9\linewidth]{sections/figure/box_ours.png}}
% \end{minipage}
%   \caption{Debiasing performance comparison of debiasing models on CelebA Dataset using box plot.}
%   \vspace{-.3cm}
% \label{fig:box_plot}
% \end{figure*}

\begin{figure*}[htbp]
\centering
\begin{minipage}[b]{.245\linewidth}
    \centering
    \subfloat[][Bias (bias amplification~\cite{Image_caption_Bias_amplification_yz}).]{\label{subfig:BA}\includegraphics[width=0.9\linewidth]{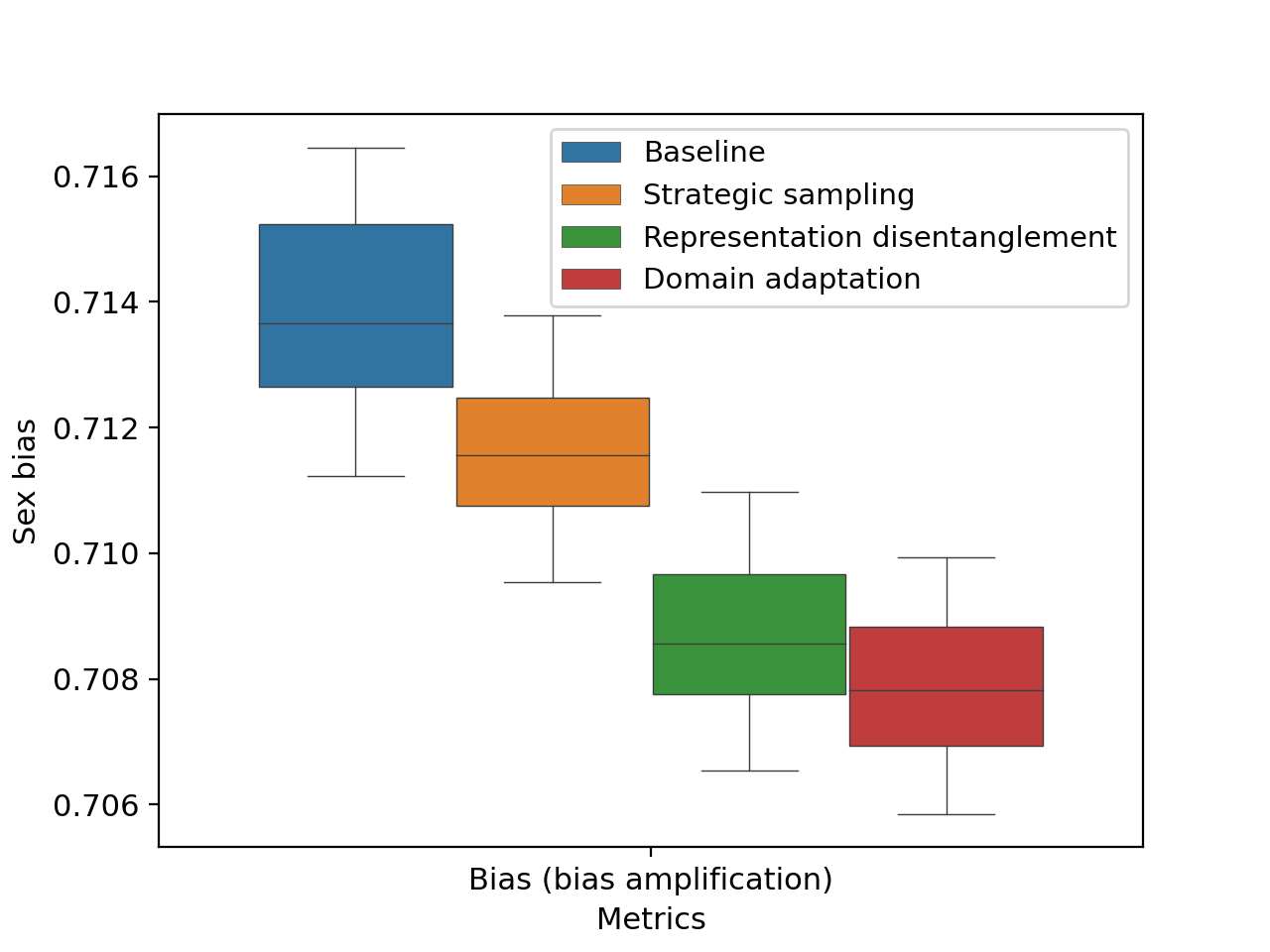}}
\end{minipage}
\begin{minipage}[b]{.245\linewidth}
    \centering
    \subfloat[][Distance correlation~\cite{correlation_distance}.]{\label{subfig:dcor}\includegraphics[width=0.9\linewidth]{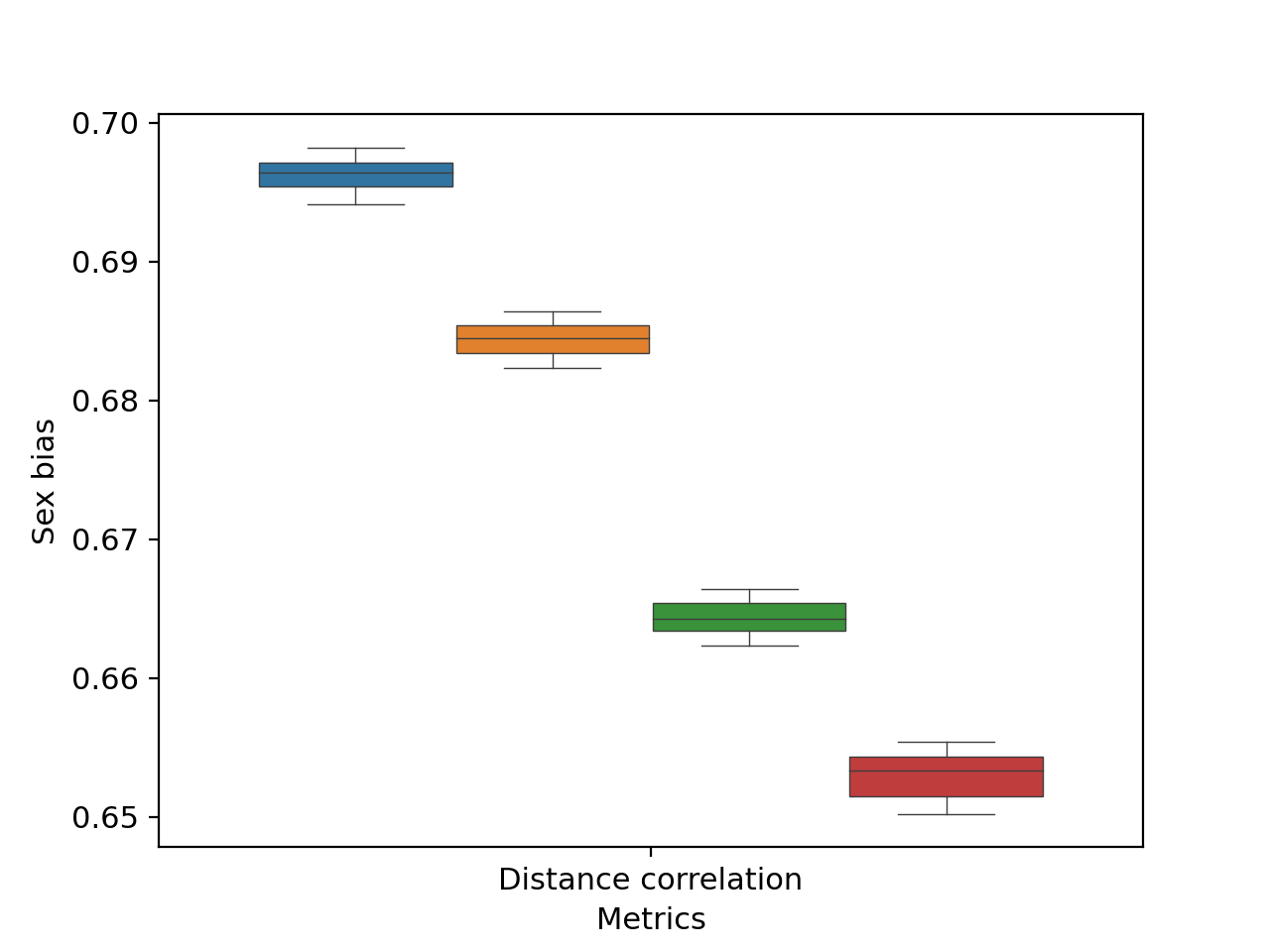}}
\end{minipage}
\begin{minipage}[b]{.245\linewidth}    
    \centering
    \subfloat[][RLB (MI baseline).]{\label{subfig:base}\includegraphics[width=0.9\linewidth]{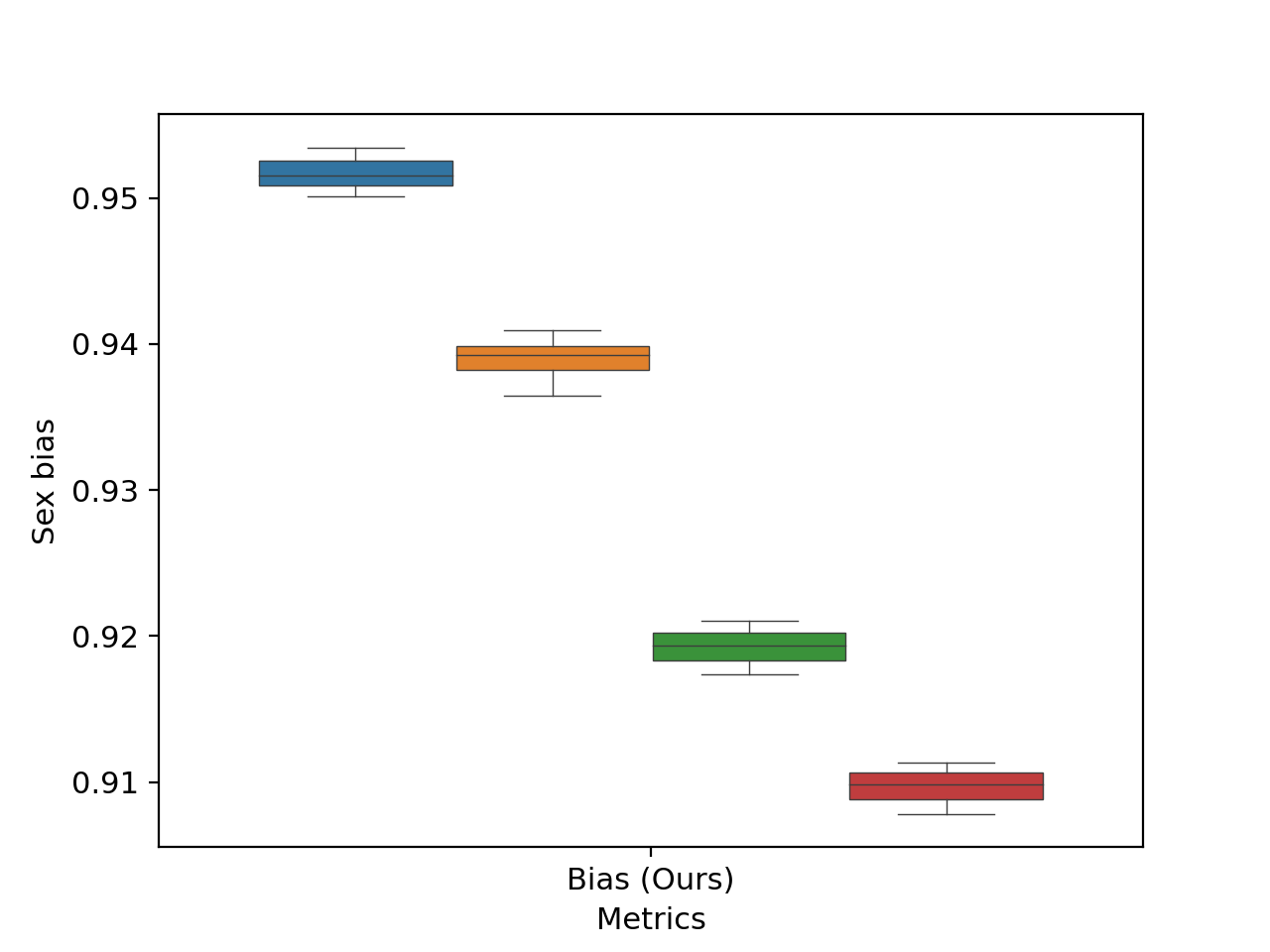}}
\end{minipage}
\begin{minipage}[b]{.245\linewidth}    
    \centering
    \subfloat[][RLB (adding mapping network).]{\label{subfig:ours}\includegraphics[width=0.9\linewidth]{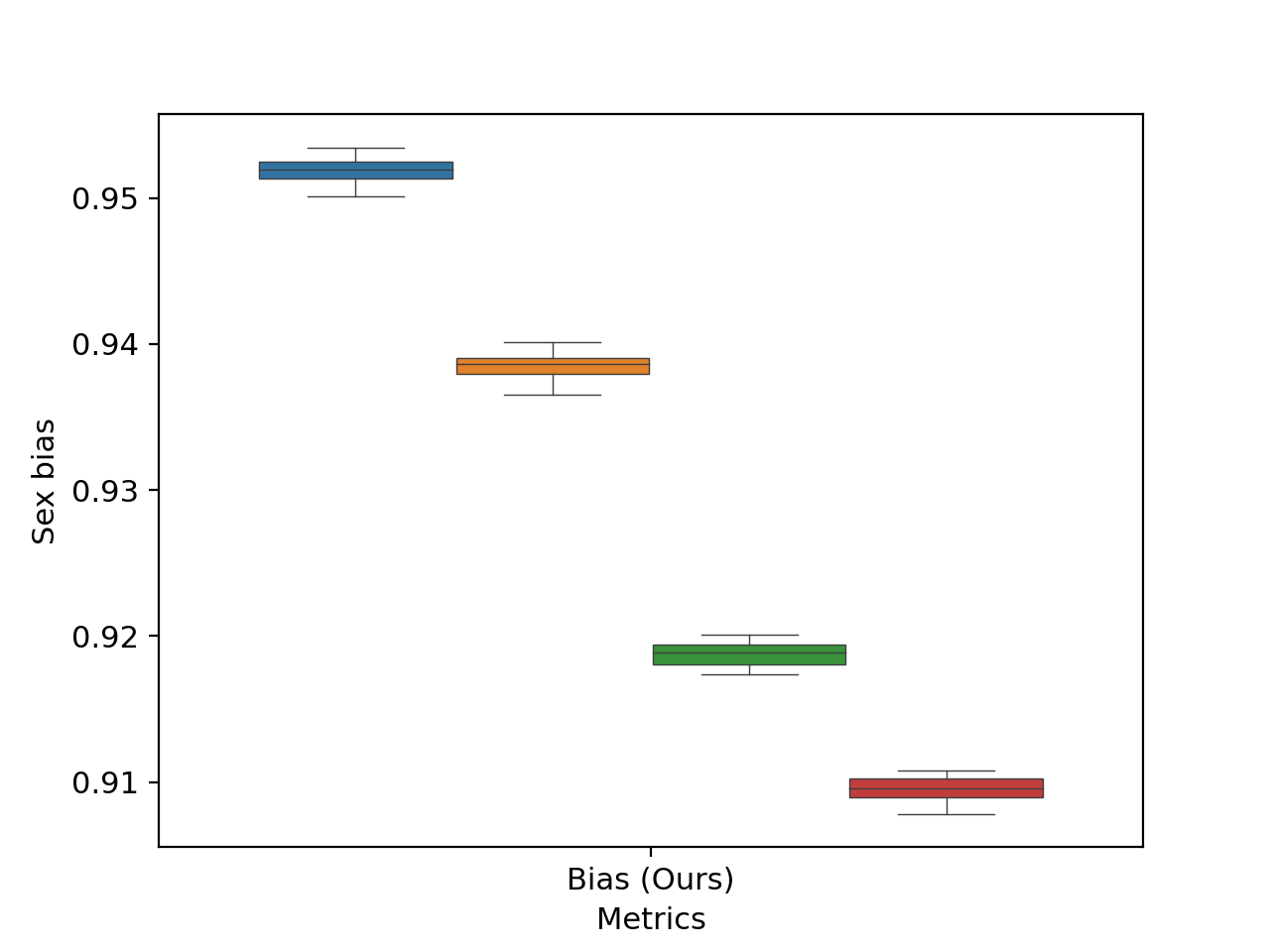}}
\end{minipage}
  \caption{Debiasing performance comparison of debiasing models on CelebA Dataset using box plots.}
  \vspace{-.28cm}
\label{fig:box_plot}
\end{figure*}

\begin{table*}[htbp]
\centering
% \small
\scriptsize
\caption{Debiasing performance comparison of debiasing models on CelebA Dataset.}
\begin{tabular}{|l|c|c|c|c|c|c|}
\hline
\multirow{2}{*}{}           & \multicolumn{3}{c|}{\textbf{mAP}}                                                                                                                           & \multicolumn{1}{c|}{\textbf{Bias (\emph{BA}~\cite{Image_caption_Bias_amplification_yz})}}   &  \multicolumn{1}{c|}{\textbf{${dcor}^2$~\cite{correlation_distance}}}                  & \multicolumn{1}{c|}{\textbf{MI / RLB (Ours)}}                \\ \cline{2-7}
                            & \textbf{Female} & \textbf{Male} & \textbf{Overall} & \textbf{Sex} & \textbf{Sex} & \textbf{Sex} \\ \hline \hline
Baseline                    & 75.8   & 72.2 & 74.3                  & 0.714               &    0.697     & \textbf{0.636 / 0.954}      \\ \hline
Strategic sampling~\cite{debias_resampling_representation_bias_colored_MNIST}      & 75.4   & 72.3 & 74.1                  & 0.712            & 0.672           & \textbf{0.628 / 0.942}       \\ \hline
Representation disentanglement~\cite{debias_adversarial_Age_sex_race_biasness, debias_adversarial_uniform_confusion_ACC_unbalanced_dataset2, debias_adversarial_forgetting_yz}     & 73.1   & 70.2 & 71.9                  & 0.708        & 0.659               & \textbf{0.620 / 0.929}       \\ \hline
Domain adaptation~\cite{debias_domain_discriminative_RFW_MI_AUC_TAR_FAR, debias_domain_discriminative_ROC_FAR}             & 74.7   & 72.5 & 73.8                  & 0.707           & 0.643            & \textbf{0.612 / 0.918}       \\ \hline
Domain independent training~\cite{debias_domain_independent_training_BA_usage} & 76.5   & 76.1 & 76.3                 & 0.702 & 0.578 & \textbf{0.553 / 0.829}       \\ \hline
\end{tabular}
  \vspace{-.25cm}
\label{tab:CelebA}
\end{table*}

\begin{table*}[htbp]
\centering
% \footnotesize
% \scriptsize
% \tiny
% \fontsize{6pt}{7pt}
\fontsize{5.6pt}{7pt}
\selectfont
\caption{Debiasing performance comparison of debiasing models on FairFace Dataset.}
\begin{tabular}{|l|c|c|c|c|c|c|c|c|c|c|c|c|c|c|c|}
\hline
\multirow{3}{*}{}           & \multicolumn{9}{c|}{\textbf{Accuracy}}                                                                                                                                                               & \multicolumn{2}{c|}{\textbf{Bias (\emph{BA}~\cite{Image_caption_Bias_amplification_yz})}}       & \multicolumn{2}{c|}{\textbf{${dcor}^2$~\cite{correlation_distance}}}           & \multicolumn{2}{c|}{\textbf{MI / RLB (Ours)}}                      \\ \cline{2-16} 
                            & \multicolumn{2}{c|}{\textbf{Black}} & \multicolumn{2}{c|}{\textbf{White}} & \multicolumn{2}{c|}{\textbf{East Asian}} & \multicolumn{2}{c|}{\textbf{Indian}} & \multirow{2}{*}{\textbf{Overall}} & \multirow{2}{*}{\textbf{Sex}} & \multirow{2}{*}{\textbf{Race}} &
                            \multirow{2}{*}{\textbf{Sex}} & \multirow{2}{*}{\textbf{Race}} & \multirow{2}{*}{\textbf{Sex}} & \multirow{2}{*}{\textbf{Race}} \\ \cline{2-9}
                            & \textbf{F}       & \textbf{M}       & \textbf{F}       & \textbf{M}       & \textbf{F}          & \textbf{M}         & \textbf{F}        & \textbf{M}       &                                   &                               &                                &                               &             & &                   \\ \hline \hline
Baseline                    & 79.3             & 79.2             & 84.2             & 88.2             & 80.4                & 80.3               & 79.1              & 81.2             & 81.3                              & 0.510                         & 0.259            & 0.573 & 0.314              & \textbf{0.480 / 0.693}               & \textbf{0.762 / 0.550}               \\ \hline
Strategic sampling~\cite{debias_resampling_representation_bias_colored_MNIST}      & 80.5             & 80.3             & 83.7             & 86.4             & 79.5                & 79.6               & 79.2              & 81.1             & 80.7                              & 0.508                         & 0.257             & 0.552 & 0.309             & \textbf{0.468 / 0.675}                & \textbf{0.757 / 0.546}                 \\ \hline
Representation disentanglement~\cite{debias_adversarial_Age_sex_race_biasness, debias_adversarial_uniform_confusion_ACC_unbalanced_dataset2, debias_adversarial_forgetting_yz}     & 78.6             & 78.5             & 82.5             & 84.5             & 78.5                & 78.6               & 78.1              & 79.3             & 78.6                              & 0.505                         & 0.255                & 0.546 & 0.286              & \textbf{0.448 / 0.646}                & \textbf{0.736 / 0.531}                 \\ \hline
Domain adaptation~\cite{debias_domain_discriminative_RFW_MI_AUC_TAR_FAR, debias_domain_discriminative_ROC_FAR}             & 80.7             & 80.6             & 83.9             & 85.6             & 79.7                & 79.8               & 79.9              & 80.1             & 79.8                              & 0.504                         & 0.255              & 0.531 & 0.281                & \textbf{0.437 / 0.631}                & \textbf{0.729 / 0.526}                 \\ \hline
Domain independent training~\cite{debias_domain_independent_training_BA_usage} & 82.5             & 82.4             & 85.3             & 87.1             & 82.5                & 82.5               & 82.3              & 82.5             & 83.5                              & 0.501                         & 0.253                      & 0.478 & 0.247        & \textbf{0.361 / 0.521}                & \textbf{0.658 / 0.475}                 \\ \hline
\end{tabular}
  \vspace{-.6cm}
\label{tab:FairFace}
\end{table*}

\noindent
\textbf{Experiment Setup.}
We simulate the degree of bias by assigning the entropy of protected attributes, such as sex entropy $H^s_i = - [P(z^f)logP(z^f) + P(z^m)logP(z^m)]$ where $P(z^f)$ is the percentage of female in the whole dataset. Next, given a source $\mathcal{S}_0$ from which the generated images are distinctive and high-quality, and a source $\mathcal{S}_1$ with four manually selected latent vectors as representatives for races, inside \emph{StyleGAN2}~\cite{styleGan2} pretrained on FFHQ~\cite{FFHQ}, the mapping network $f$ produces $e_0^{(i)}$ and $e_1^{(j)}$ from $u_0^{(i)} \in \mathcal{S}_0$ and $u_1^{(j)} \in \mathcal{S}_1$, and the synthesis network $p$ generates an image $I_{ij}$ by taking $e_0^{(i)}$ at coarse spatial resolution $(4^2 - 8^2)$ to bring high-level appearances (hair style, face shape) from $\mathcal{S}_0$, and $e_1^{(j)}$ at fine spatial resolution $(16^2 - 1024^2)$ to obtain racial appearances (colors of eyes, hair, skin) from $\mathcal{S}_1$. Further, we generate a specific dataset according to the preset entropy. Finally, we train a ResNet-50~\cite{ResNet50} on generated datasets with different sex entropy $H^g_i$ or race entropy $H^r_i$ to construct learned representation space $\mathcal{R}$ for evaluating our representation-level \emph{independent} bias assessment on the balanced testing set with $H^g_i = 0.693$ and $H^r_i = 1.386$ at the balance point. We use skin color as a proxy to race in this experiment, same as Pilot Parliaments Benchmark (PPB)~\cite{sex_race_PPB}. The end-to-end three-stage pipeline is shown in~\cref{fig:stylegan2_whole_pipeline}.

\noindent
\textbf{Results.}
As shown in~\cref{fig:generated_dataset_sample}, benefiting from \textit{mapping network} to reduce \textit{feature entanglement}, the generated images in different races maintain the similar appearance with different skin tone, which mitigates interference of appearance. A few observations can be drawn from the~\cref{fig:entropy}. First, RLB declines as the entropy increases. According to the definition of entropy, larger entropy implies a more balanced dataset, and therefore, RLB is consistent with degree of imbalance. Second, compared to sex entropy, race entropy has a stronger influence on RLB. 

\subsection{Comparison With Debiased Models}
\label{Comparison_debiasing_models}
Inspired by~\cite{debias_domain_independent_training_BA_usage}, we compare bias assessment metrics on four mainstream families of debiasing methods --- (1) strategic sampling, (2) representation disentanglement, (3) domain adaptation and (4) domain independent training, with BA~\cite{Image_caption_Bias_amplification_yz} and ${dcor}^2$~\cite{correlation_distance}. The ResNet-50~\cite{ResNet50} pre-trained on ImageNet~\cite{imagenet} (as baseline model) is used to predict attributes. We assess RLB of sex on CelebA dataset and both sex and race on FairFace dataset. Mean average precision (mAP) across cohorts is also presented as metric comparison for this multi-label classification.   

\noindent
\textbf{Results.}
In order to illustrate the ability to capture model bias, the degree of imbalance of testing dataset is kept same, \ie sex entropy of CelebA dataset is $0.667$, sex entropy and race entropy of FairFace dataset is $0.693$ and $1.386$. Besides, mutual information estimation (MI) corresponding with RLB is separably presented as ablation study of bias assessment with and without entropy. In \cref{tab:CelebA} and \cref{tab:FairFace}, debiasing models are compared across rows and metrics are compared across columns. The results show that domain-independent training~\cite{debias_domain_independent_training_BA_usage} performs the best with the most balanced mAP across all cohorts. In order to present variation and mean of different metrics in a more straightforward way, we conduct experiments with 50 different random seeds and calculate statistics of different metrics for sex bias in CelebA dataset~\cite{CelebA} using box plot, as shown in \cref{fig:box_plot}. Comparing \cref{subfig:BA} with \cref{subfig:ours}, we find that the confused four groups under BA~\cite{Image_caption_Bias_amplification_yz} are clearly distinguished under RLB due to a larger range which can be used to stratify different degree of bias from more models without aliasing. Furthermore, comparing with ${dcor}^2$ in \cref{subfig:dcor}, RLB assesses sex bias with small variation. Also, comparing \cref{subfig:base} and \cref{subfig:ours}, RLB (adding mapping network) yields smaller variation and more robustness than MI baseline. Theoretically, different metrics for demographic bias construct a \emph{metric space} rather than a \emph{normed space} since there is no definition of \emph{zero point}. Furthermore, in the absence of standard unit of bias and proper conversion between different metrics, we need to consider the absolute value instead of the relative value and the advantages (clear discrepancy and small variation) of RLB demonstrate a better precision. 

%  or define a origin point for metric comparison. two different metric space containing distance, distance plus zero point equal to norm, then we can compare so that we need normalization (upper bound or lower bound), normed vector spaces, A norm induces a distance, a metric (a notion of distance) d(X,Y) = ||X-Y||, norm of X can be simply regarded as distance between X to zero point

% Further, compared to the harsh criterion (DP), mAP across cohorts is a soft metric so that it is used as the reference in Tab.\red{2}\&\red{3}.

% \begin{table}[htb]
% \centering
% % \footnotesize
% \scriptsize
% % \tiny
% \caption{Comparison between correlations established by predictor and mutual information.}
% \begin{tabular}{|c|c|c|c|c|c|}
% \hline
%             & \textbf{Normal} & \bm{$R_S$} & \bm{$R_G$}    & \bm{$Z_S$} & \bm{$Z_G$} \\ 
%             & \textbf{Correlation} & & & & \\ \hline
% Accuracy & 99.7 & 63.8 & 99.9 & 92.3 & 99.8 \\ \hline
% \emph{Logits} Loss & 0.054     & 0.412     & 0.23e-05 & 0.179  & 0.026  \\ \hline
% \textbf{Bias (Ours)} & \textbf{0.874}          & \textbf{2.42e-05}     & \textbf{0.34e-05} & \textbf{1.28e-05}  & \textbf{1.09e-05}  \\ \hline
% \end{tabular}
% \label{tab:synthesized_representation2}
% \end{table}

% \subfile{sections/05_comparison}
\section{Conclusion}
We present a bias assessment metric to assess demographic bias in face recognition at the representation level and empirically demonstrate that RLB reflects the bias issues induced from imbalanced datasets and biased models. Our results show that the conclusions of previous work that use mAP across cohorts, BA and ${dcor}^2$ show large variation and may produce contradictory bias assessment scores when comparing more debiasing models since these metrics have not yielded clear discrepancy. Furthermore, the conclusions of prior work that use \logits~loss to evaluate debiasing performance may be inaccurate since spurious correlations may lead to inaccurate \logits-level metrics. On the other hand, our \emph{independent} representation-level bias can be not only used to evaluate the overall performance for bias mitigation, but also used to detect bias inside debiased models, which allows a more flexible and wider-range usage for studying bias in classification models.
% \clearpage

% \addtolength{\textheight}{-3cm} 

% \section{ACKNOWLEDGMENTS}

% The authors gratefully acknowledge the contribution of reviewers' comments, etc. (if desired). Put sponsor acknowledgments in the unnumbered footnote on the first page.

{
\small
% \tiny
\bibliographystyle{ieee}
\bibliography{reference}
}

\end{document}